\documentclass[runningheads]{llncs}
\usepackage{graphicx}

\usepackage{tikz}
\usepackage{comment} 
\usepackage{amsmath,amssymb} 
\usepackage{color}
\usepackage{multirow}
\usepackage{float}
\usepackage{caption}
\usepackage[ruled,vlined,linesnumbered]{algorithm2e}

\begin{document}
\pagestyle{headings}
\mainmatter
\def\ECCVSubNumber{444}  

\title{Meshing Point Clouds with Predicted Intrinsic-Extrinsic Ratio Guidance}  

\titlerunning{Meshing Point Clouds with Predicted Intrinsic-Extrinsic Ratio Guidance}

\author{Minghua Liu \and
Xiaoshuai Zhang \and
Hao Su}
\authorrunning{M. Liu et al.}

\institute{University of California, San Diego \\
\email{\{minghua,zxs,haosu\}@ucsd.edu}}

\maketitle

\begin{abstract}
We are interested in reconstructing the mesh representation of object surfaces from point clouds. Surface reconstruction is a prerequisite for downstream applications such as rendering, collision avoidance for planning, animation, etc. However, the task is challenging if the input point cloud has a low resolution, which is common in real-world scenarios (e.g., from LiDAR or Kinect sensors). Existing learning-based mesh generative methods mostly predict the surface by first building a shape embedding that is at the whole object level, a design that causes issues in generating fine-grained details and generalizing to unseen categories. Instead, we propose to leverage the input point cloud as much as possible, by only adding connectivity information to existing points. Particularly, we predict which triplets of points should form faces. Our key innovation is a surrogate of local connectivity, calculated by comparing the intrinsic/extrinsic metrics. We learn to predict this surrogate using a deep point cloud network and then feed it to an efficient post-processing module for high-quality mesh generation. We demonstrate that our method can not only preserve details, handle ambiguous structures, but also possess strong generalizability to unseen categories by experiments on synthetic and real data. The code is available at https://github.com/Colin97/Point2Mesh.
\end{abstract}

\section{Introduction}
Among various 3D representations (e.g., polygonal meshes, voxels, point clouds, multi-view 2D images, part-based primitives, and implicit field functions), polygonal meshes capture the geometric details of the shape in an efficient way, which prevents high memory footprints and artifacts caused by discretization. Reconstructing high-quality 3D meshes from point clouds thus has been studied for quite a long time and serves as a prerequisite for numerous real-world applications, including autonomous driving, augmented reality, and robotics. 

Despite its long history, the mesh reconstruction problem remains unresolved. Traditional methods~\cite{kazhdan2006poisson,bernardini1999ball,lorensen1987marching} typically reconstruct the mesh either by explicitly connecting the points or implicitly approximating the surface, both of which resort to local geometric hints. Without reasoning about the shape, traditional methods may be hard to handle the ambiguous structures when the resolution of the input point cloud is limited. For example, the ambiguous structures may include thin structures consisting of two very close surfaces, independent but spatially adjacent parts, and corners. Traditional methods tend to produce distortion or connect independent parts incorrectly when facing these structures. However, the reconstruction of these fine-grained structures may be essential for many downstream applications such as robotics grasping which needs an accurate understanding of part-level mobility. Moreover, traditional methods are typically sensitive to hyper-parameters. For most of these methods, a dedicated parameter-tuning is required for each input, making batch processing of point clouds impractical.  

With the rapid development of 3D deep learning and the availability of large-scale 3D datasets, people tend to learn geometric or semantic priors from data. Unlike 2D images and 3D voxels, polygon meshes is an irregular geometric representation, which prevents it from being generated by the neural network directly. However, there are still lots of attempts to explore the neural-network-compatible representations for mesh generation, including template meshes with deformation~\cite{wang2018pixel2mesh,groueix20183d,gao2019sdm,pan2019deep,liu2019soft,kanazawa2018learning,gkioxari2019mesh,wen2019pixel2mesh++}, 2D squares with folding~\cite{yang2017foldingnet,groueix2018papier,deprelle2019learning}, primitives with assembly~\cite{chen2019bsp,tulsiani2017learning,sharma2018csgnet}, implicit field function~\cite{park2019deepsdf,chen2019learning,genova2019learning,mescheder2019occupancy}, and meshlets with optimization~\cite{badki2020meshlet,williams2019deep}. Existing learning-based methods typically follow the ``encoder-decoder'' paradigm. The limited capability of the network prevents existing methods from generating fine-grained structures and details. Also, since most existing methods learn the priors at the object level, they tend to memorize the overall shapes and typically cannot generalize to unseen categories.

To this end, we propose a novel method that reconstructs meshes from point clouds by leveraging the intermediate representation of triangle faces. Unlike existing methods, our method fully utilizes the input point clouds, which are on the ground truth surface in most cases, and then estimate the local connectivity with the help of learned guidance. More specifically, we first propose a set of candidate triangle faces, which could be the elements of the reconstructed mesh, by constructing a $k$-nearest neighbor ($k$-NN) graph on the input point cloud. We then utilize the neural network to filter out the incorrect candidates and provide cues for sorting the remaining candidates. We find that the ratio of geodesic distance (intrinsic metric) and Euclidean distance (extrinsic metric) between two vertices may provide strong cues for inferring the connectivity and can naturally serve as the supervision for the candidate classification task. Since there are multiple ways to triangulate a surface, we only filter out those candidates that should never appear in the reconstructed mesh, such as the candidates linking two independent parts. A greedy post-processing algorithm is then used to sort all the remaining candidates and merge them into the final mesh.

We demonstrate that our algorithm can preserve fine-grained details and handle ambiguous structures with the help of learned intrinsic-extrinsic guidance. Since our method reconstructs meshes by estimating local connectivity, which relies mainly on the local geometric information, it can well generalize to unseen categories. In experiments on the ShapeNet dataset, our method outperforms both the existing traditional methods and learning-based mesh generative methods with regard to all commonly used metrics, including the F-score, Chamfer distance, and normal consistency. We also provide extensive ablation studies on different sampling densities, sampling strategies, noisy levels, and real scans to demonstrate our generalizability and robustness. 

\section{Related Work}
3D mesh reconstruction is a core problem for many applications. Yet despite its long history, the problem is still far from being solved. In this section, we review the existing methods and the remaining difficulties of the problem.

\subsection{Traditional Mesh Reconstruction}
Traditional mesh reconstruction methods mainly include two paradigms: explicit reconstruction and implicit reconstruction. 

Explicit reconstruction methods, such as ball-pivoting algorithm (BPA)~\cite{bernardini1999ball}, Delaunay triangulation~\cite{boissonnat1993three}, alpha shapes~\cite{edelsbrunner1994three}, and zippering~\cite{turk1994zippered}, resort to the local surface connectivity estimation and connect the sampled points directly by triangles. For example, the principle of BPA is simple: three points form a triangle if a ball of a user-specified radius touches them without containing any other point. However, the radius of BPA matters a lot: a small radius can lead to holes while a large radius may cause incorrect connections. Although there are some following works trying to utilize multiple radii~\cite{digne2014analysis}, they still fail to handle ambiguous structures well. 

Implicit reconstruction methods~\cite{kazhdan2006poisson,kazhdan2013screened,guennebaud2008dynamic,oztireli2009feature,carr2001reconstruction,hoppe1992surface} try to find a field function (e.g., signed distance function) approximating the point cloud and then employ the marching cube algorithm~\cite{lorensen1987marching} to extract the iso-surface of the field function.  For example, Poisson surface reconstruction (PSR)~\cite{kazhdan2006poisson,kazhdan2013screened} reconstructs the surface by solving a Poisson problem for the oriented points. However, solving large-scale equations is time-consuming. Also, it is difficult for traditional algorithms to determine the consistent direction of the normals based only on the coordinates of the point cloud. Without correct vertex normal directions, PSR tends to generate poor results. Moreover, implicit reconstruction methods utilize marching cube~\cite{lorensen1987marching} to generate the mesh, which may lead to expensive voxelization and the artifacts caused by the discretization. 

Under the limited resolution, ambiguities of the input point cloud require the integration of strong geometric or semantic priors about our 3D world. With such priors and reasoning, our learning-based methods are expected to handle those ambiguous structures and avoid results with distortion and artifacts. In addition, traditional algorithms heavily rely on selecting a set of proper hyper-parameters, which may require a case-by-case parameter tuning, while our learning-based algorithm should be applied to all cases adaptively and thus enable automatic batch processing. 

\subsection{Learning-based mesh generation}

The recent success of 3D deep learning~\cite{qi2017pointnet,qi2017pointnet++} and the availability of large 3D datasets~\cite{chang2015shapenet,mo2019partnet} nourish the tasks of 3D analysis and 3D synthesis. However, unlike 2D images, 3D polygon meshes are irregular geometric formats and are difficult to be directly generated from the neural networks. Existing learning-based mesh generative methods mainly follow five paradigms: deformation-based methods~\cite{wang2018pixel2mesh,groueix20183d,gao2019sdm,pan2019deep,liu2019soft,kanazawa2018learning,gkioxari2019mesh,wen2019pixel2mesh++}, folding-based methods~\cite{yang2017foldingnet,groueix2018papier,deprelle2019learning}, primitive-based methods~\cite{chen2019bsp,tulsiani2017learning,sharma2018csgnet}, optimization-based methods~\cite{badki2020meshlet,williams2019deep}, and implicit-field-function-based methods~\cite{park2019deepsdf,chen2019learning,genova2019learning,mescheder2019occupancy,liao2018deep}.

Deformation-based methods resort to deform a template mesh (e.g., a sphere mesh) into the desired shape. However, since they only deform the position of the vertices without changing the connectivity, the topology of the template mesh may restrict the methods from generating shapes of a specific topology. Folding-based methods learn a set of mappings from 2D squares to 3D patches, which are then used to form the mesh. Primitive-based methods utilize a set of primitives (e.g., planes and convex patches) to form the final mesh, and learn the parameters of the primitives. The simplicity of the primitives may prevent the methods from generating fine-grained details. As for the implicit-field-function-based methods, they employ neural networks to learn an implicit field function and then utilize marching cube algorithms~\cite{lorensen1987marching} to extract the iso-surface, and thus face similar problems as the traditional implicit reconstruction methods. There are also some recent optimization-based methods, which either utilize a deep neural network as local geometric prior~\cite{williams2019deep} or learn some local shape priors from the data~\cite{badki2020meshlet}. They formulate mesh reconstruction as an optimization problem and are thus computational expensive. 

Most existing learning-based methods do not make full use of the input point cloud that processing should be grounded upon. Although they may be able to generate the coarse-grained shapes, they may fail to capture some of the structures and details. Also, most existing methods learn the priors at the object level, which makes them category-specific, and even sensitive to the pose of the object. In contrast, our method will be fully based on the grounded point clouds to preserve all the structures and generate fine-grained details. Moreover, our method relies on local priors and can thus generalize to unseen categories.
 
\section{Method}
Given a 3D point cloud $P = \{(x_i, y_i, z_i)\}$, we aim to reconstruct a polygon mesh, which consists of a vertex set and a face set, approximating the underlying surface. Unlike existing learning-based mesh generative methods, we fully utilize the grounded input point cloud and let $P$ serve as the vertex set of the reconstructed mesh, since they are usually on or near the surface of the shape and provide lots of cues for the structures and details. We then reconstruct the mesh by predicting the local connectivity between the vertices. Before presenting our reconstruction method, we would like to introduce a motivating remeshing algorithm, where the ground truth mesh is known. We then extend the remeshing algorithm to mesh reconstruction by introducing a neural network module.

\subsection{A Motivating Remeshing Algorithm}
\begin{figure}[t]
\centering
\includegraphics[width=0.9\linewidth]{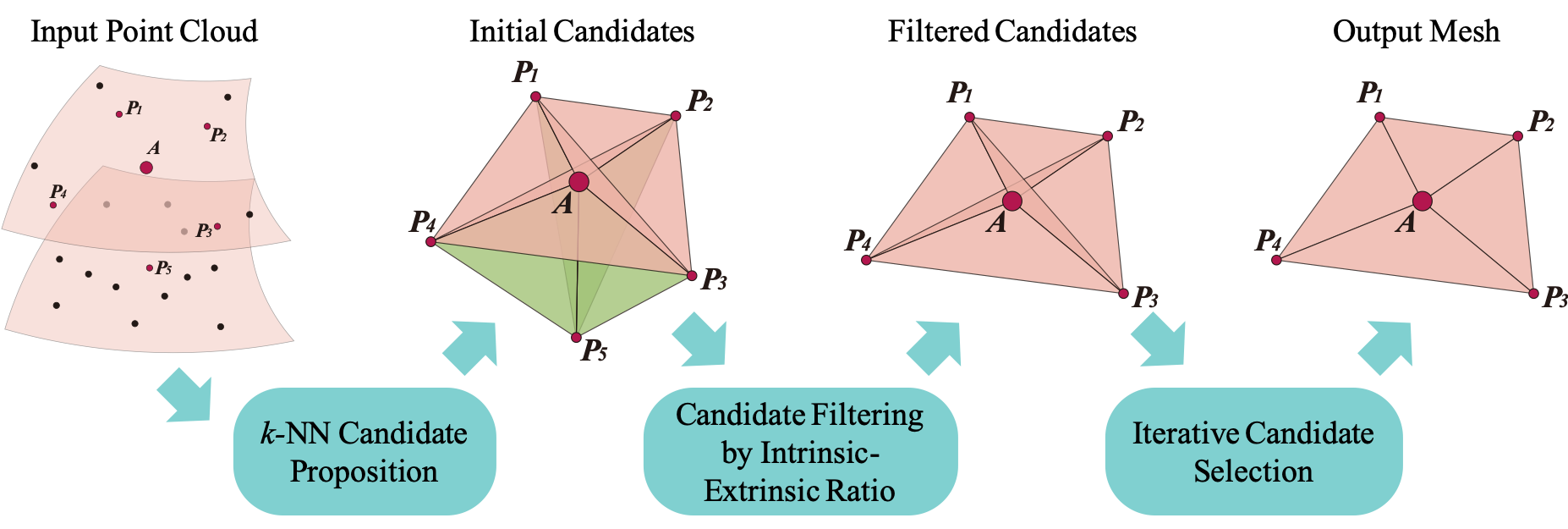}
\caption{Our remeshing pipeline from a local view. The candidate triangles near the underlying surfaces are colored in pink, while others are colored in green.}
\label{fig:remeshpip}
\end{figure}

Given a reference triangle mesh $M_R$ and a point cloud $P$ sampled on $M_R$ as input, the remeshing algorithm aims to generate a new mesh $M_N$ whose vertices come from the point cloud $P$. As shown in Fig.~\ref{fig:remeshpip}, the algorithm first proposes a set of candidate triangle faces and then uses a subset of candidate triangles to form the mesh $M_N$.

\subsubsection{Candidate Proposition} Since each vertex should only connect to its neighbors on the surface, the algorithm first constructs a $k$-nearest neighbor ($k$-NN) graph for each of the points in $P$ based on the Euclidean distance, and then each vertex form candidate triangle faces with each two of its $k$-NN neighbors. We expect the union of candidate triangles to cover the whole surface of $M_R$, which means for an even surface with uniformly distributed vertices, a small $k$ would be enough, while for complex surfaces and nonuniformly distributed vertices, we may need a larger $k$. However, we could select a $k$ that is large enough to cover most practical cases. Among all the candidates, some of them are on or near the surface of mesh $M_R$, and we denote them as \emph{correct candidates}, while the others are away from the surface and are denoted as \emph{incorrect candidates}. The incorrect candidates may appear in areas such as (a) thin structures consisting of two very close surfaces, (b) independent but spatially adjacent parts, and (c) surfaces with large curvature. We would like to filter out the incorrect candidates, and use some of the remaining candidate faces to form the mesh $M_N$. 

\begin{figure}[t]
\centering
\includegraphics[width=\linewidth]{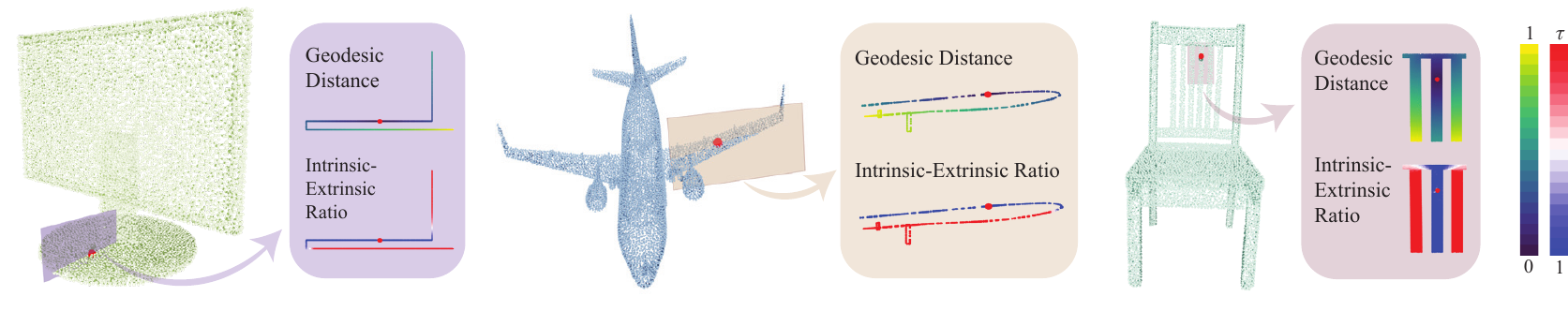}
\caption{In each example, we sample a slice from the input point cloud, and demonstrate the geodesic distance and the intrinsic-extrinsic ratio (IER) to the key point (marked in red). }
\label{fig:geo}
\end{figure}

\subsubsection{Candidate Filtering} Since the reference mesh $M_R$ is given, we can calculate the geodesic distance between two vertices, which is defined to be the length of the shortest path over the surface of $M_R$. As the intrinsic metric of the surface manifold, geodesic distance provides strong cues for inferring the connectivity between two vertices. The geodesic distance could be inconsistent with the Euclidean distance (extrinsic metric). However, in a small neighborhood, if the geodesic distance between two vertices is equal or close to their Euclidean distance, the connecting line between them is likely to be on or close to the surface of mesh $M_R$. We thus define the intrinsic-extrinsic ratio (IER) for a pair of vertices $u,v$ as:
\begin{equation}
    \text{IER}(u, v) = \frac{d_G(u,v)}{d_E(u,v)},
\label{equ:ier}
\end{equation}
where $d_G$ and $d_E$ indicate the geodesic distance and the Euclidean distance respectively. Fig.~\ref{fig:geo} shows some examples of how geodesic distance and the IER to a key point change within a slice. The key point should not be connected with the points in the red region where IER is much higher than 1. The cases in Fig.~\ref{fig:geo} demonstrate that IER can effectively handle corners (see display), thin structures (see jet), and adjacent parts (see chair). We thus propose to employ the IER to filter out the incorrect triangle candidates. For a triangle face with vertices $u$, $v$, and $w$, the IER is extended to be:
\begin{equation}
\text{IER}(u, v, w) = \frac{d_G(u,v) + d_G(u, w) + d_G(v, w)}{d_E(u,v) + d_E(u, w) + d_E(v, w)}.
\label{equ:iert}
\end{equation}
With the definition above, we filter out candidates whose IER is greater than $\tau$, and $\tau > 1$ is a preset threshold. After filtering out the incorrect candidates, the remaining candidates should be on or near the surface of $M_R$. 

\subsubsection{Sort and Merge} Since there is no canonical way to triangulate a surface, we sort the remaining candidate triangles and merge them into mesh $M_N$ in a greedy way. Specifically, we prefer triangles that are closer to the surface of $M_R$. Also, we prefer candidate triangles with three short edges, which typically correspond to the equilateral triangles when the input point cloud is uniformly distributed. We thus sort the remaining candidates with respect to their distance to $M_R$ and the length of their longest edge. Specifically, we first divide the remaining candidate triangles into $l$ bins based on the distance to $M_R$, and then sort them in each bin according to the length of their longest edge. After sorting the triangles, we visit each candidate one by one and add the candidate into the mesh $M_N$ if it satisfies two constraints. Specifically, the new candidate face should not intersect with the previously added faces. Also, $M_N$ should not contain non-manifold edges after adding the new candidate. That is, $M_N$ should not contain an edge that has more than two incident faces. If both constraints hold, the candidate will be added into $M_N$; otherwise, it will be discarded. After visiting all the candidates, we get the final mesh $M_N$. 

\begin{figure}[t]
\centering
\includegraphics[width=0.8\linewidth]{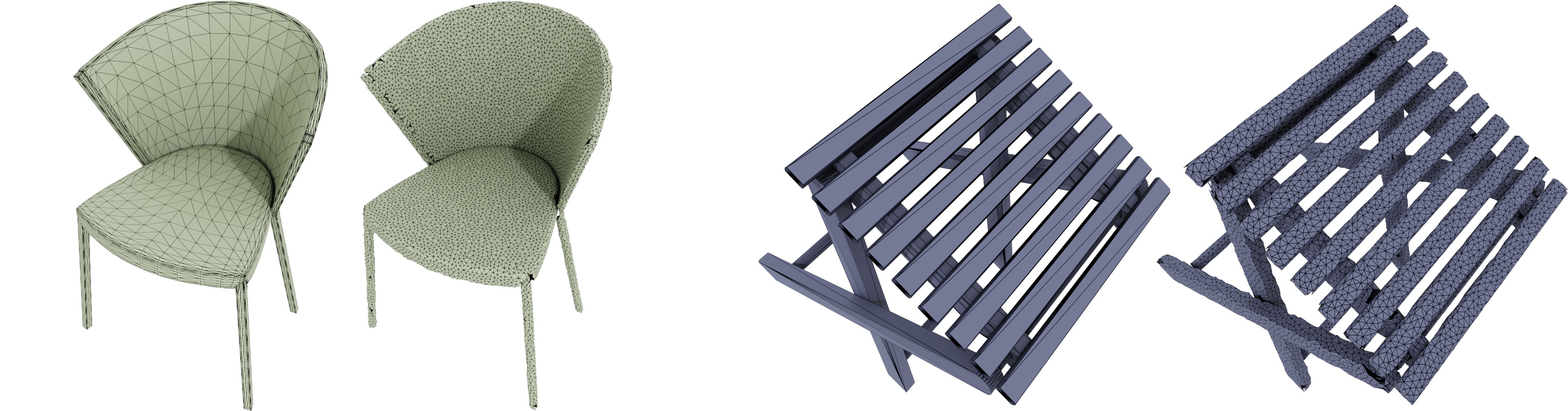}
\caption{In each pair, the left one is the original reference mesh ($M_R$), and the right one is the result of our remeshing algorithm ($M_N$).}
\label{fig:remeshing}
\end{figure}

Please refer to our supplementary materials for the pseudo code of the remeshing algorithm. Some remeshing results of the algorithm are shown in Fig.~\ref{fig:remeshing}.

\subsection{From Remeshing to Reconstruction}
\begin{figure}[t]
\centering
\includegraphics[width=\linewidth]{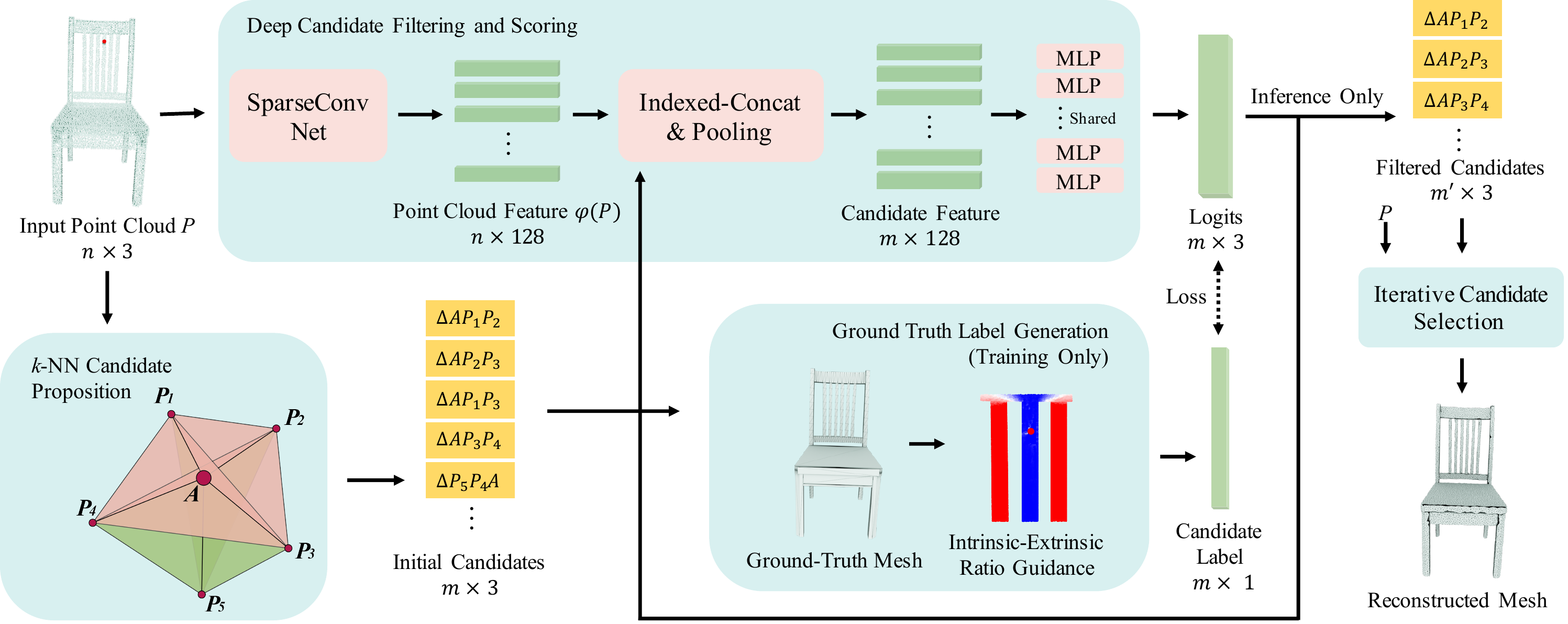}
\caption{Full pipeline of our reconstruction algorithm: given a point cloud as input, we first propose a set of candidate triangles. During training, the network is trained to classify the candidate triangles with the supervision of intrinsic-extrinsic ratio. During inference, the predicted label is used to filter out and score the remaining candidate triangles, which are then merged into the output mesh by our iterative selection algorithm.}
\label{fig:framework}
\end{figure}

In the mesh reconstruction setting, we are only given a point cloud $P$ as input and aim to reconstruct the mesh $M_N$. As shown in Fig.~\ref{fig:framework}, we follow the remeshing algorithm to construct a $k$-NN graph on $P$ and propose $m$ candidate triangle faces. Unlike in the remeshing algorithm, the reference mesh is not available, we thus cannot directly calculate the intrinsic-extrinsic ratio to serve as the guidance. At this point, the neural network may be helpful for estimating the local connectivity and geometry of the input point cloud. We thus resort to the neural network to filter out the incorrect candidate triangles and provide cues for sorting the remaining candidates. 

In training, we have ground truth mesh and we follow the above remeshing algorithm to generate a label for each candidate, which serves as dense supervision for the candidate classification network. Specifically, the candidates are divided into $l+1$ categories. The incorrect candidates with $\text{IER} \geq \tau$ are in category 0. The remaining candidates are near the ground truth surface and are divided into $l$ categories according to their distances to the surface. Empirically, we set $l = 2$ in all our experiments, which means candidates that are very close to the ground truth surface are in category 1, and other correct ones are in category 2. The network is thus trained to predict a 3-class label for each candidate.

As shown in Fig.~\ref{fig:framework}, our network consists of several parts. We first utilize SparseConvNet~\cite{graham20183d}, which are designed to process spatially-sparse data with convolutional operations, for point cloud feature extraction. Specifically, it maps the input point cloud into a feature set $\left\{\varphi\left(p_{i}\right) | p_{i} \in P\right\}$, where $\varphi\left(p_{i}\right) \in \mathbb{R}^{3+C}$ is a concatenation of the $xyz$ coordinates and the $C$ dimensional embedding of a point $p_i$. The three latent features $\varphi\left(u\right)$, $\varphi\left(v\right)$, and $\varphi\left(w\right)$ are then concatenated together for every candidate triangle face with vertices $u$, $v$ and $w$.  A symmetry function (e.g., a max-pooling layer) then takes the concatenated feature as input to aggregates the information from the three vertices. The resulting tensors are fed into a shared weight multiple layer perceptron (MLP) to predict the final label for each triangle. The convolutional operation and the shared weight MLP are designed to inspire the learning of generalizable local priors across different parts and shapes.

During inference, we utilize the predicted labels to filter out the incorrect candidates and then sort the remaining candidates according to their labels and the length of their longest edges. We finally merge them into the output mesh through a greedy post-processing, as in the remeshing algorithm.

Although we utilize the ratio between the geodesic distance and the Euclidean distance (IER) to determine the label of the candidate triangles, the neural network does not need to regress the geodesic distances directly, since it could learn to utilize local geometric and semantic cues to recognize those incorrect triangles. In our experiments, the network achieves high accuracy classifying the candidate triangles, and thus enables high-quality mesh reconstruction. 
Please refer to the supplementary materials for the experiment of directly inferring the geodesic distance, which produces less effective results. 

We find our model can transfer well to unseen categories. Since the estimation of local connectivity relies more on local inductive biases, which encourages the generalizability. Similar phenomenons are also observed in \cite{liu2020neural,badki2020meshlet,luo2020learning}. Although our method proposes $O(k^2n)$ candidate triangles, the candidate classification can be processed in batch. The total inference time for a single point cloud with $12,800$ points is typically less than 10 seconds.

\section{Experiments}
\subsection{Data Generation and Network Training} 
To evaluate our method, we sampled 23,108 synthetic CAD models, which cover eight categories, from the ShapeNet dataset~\cite{chang2015shapenet}. All the models are normalized to the origin of the canonical frame with a diameter of 1. Since there is no watertight or manifold guarantee for the ShapeNet meshes, we pre-clean the meshes to facilitate the calculation of geodesic distance. Specifically, we merge the vertices that are within a small distance of 0.001, remove all the duplicate faces, and split the edges that go through the non-endpoint vertices. For each model, we then utilize the Poisson-disk sampling~\cite{corsini2012efficient} to uniformly sample 10,000 $\sim$ 12,800 points on the mesh surface. The points are then unified into the size of 12,800 by randomly replicating, and serve as the input point cloud. For each vertex, we construct a $k$-NN graph ($k=50$). We follow the idea of the MMP algorithm~\cite{surazhsky2005fast} to calculate the exact geodesic distance between each vertex and its small neighborhood over the mesh surface. For calculating the Euclidean distance between a candidate triangle and the ground truth mesh, we randomly sample 10 points on the candidate face and average the distances of the sampled points to the ground truth mesh surface. To filter out the incorrect candidates, we empirically set $\tau$ to be $1.3$ in our experiments. The correct candidates are then divided into two categories according to their distances to the mesh surface with a threshold of 0.005.

We reserve 3,146 models for testing, and the rest is used for training. All the 8 categories are trained together in a single model. The resolution of the SparseConvNet is set to be 150.  In each training iteration, we randomly sample 25,000 candidate triangles per shape. For the following evaluations, our models were trained on a single Nvidia 2080Ti GPU for 50 epochs with a batch size of 24. The Adam optimizer was used. The initial learning rate was set to 1e-3 and decayed by 0.7 per 5 epochs.

\subsection{Comparison with Existing Methods}
\begin{table}[h]
  \centering
  \caption{Quantitative results on the ShapeNet test set: F-score with two different thresholds, Chamfer distance, and normal consistency score.}
  \tiny
    \begin{tabular}{c|cccccccc|cccccccc}
    \hline
          & \multicolumn{8}{c|}{F-score ($\mu$) $\uparrow$ }    & \multicolumn{8}{c}{F-score (2 $\mu$) $\uparrow$} \\
    \hline
        \multicolumn{1}{c|}{category} & \tiny{PSR}   & \tiny{MC}    & \tiny{BPA}   & \tiny{ATLAS} & \tiny{DMC} & \tiny{DSDF}  & \tiny{DGP}   & \tiny{OURS}  & \tiny{PSR}   & \tiny{MC}    & \tiny{BPA}   & \tiny{ATLAS} &  \tiny{DMC} & \tiny{DSDF}  & \tiny{DGP}   & \tiny{OURS} \\ \hline
            display & 0.468 & 0.495 & 0.834 & 0.071 & 0.108 & 0.632 & 0.417 & \textbf{0.903} & 0.666 & 0.669 & 0.929 & 0.179 & 0.246 & 0.787 & 0.607 & \textbf{0.975} \\
    lamp  & 0.455 & 0.518 & 0.826 & 0.029 & 0.047 & 0.268 & 0.405 & \textbf{0.855} & 0.648 & 0.681 & 0.934 & 0.077 & 0.113 & 0.478 & 0.662 & \textbf{0.951} \\
    airplane & 0.415 & 0.442 & 0.788 & 0.070 & 0.050 & 0.350 & 0.249 & \textbf{0.844} & 0.619 & 0.639 & 0.914 & 0.179 & 0.289 & 0.566 & 0.515 & \textbf{0.946} \\
    cabinet & 0.392 & 0.392 & 0.553 & 0.077 & 0.154 & 0.573 & 0.513 & \textbf{0.860} & 0.598 & 0.591 & 0.706 & 0.195 & 0.128 & 0.694 & 0.738 & \textbf{0.946} \\
    vessel & 0.415 & 0.466 & 0.789 & 0.058 & 0.055 & 0.323 & 0.387 & \textbf{0.862} & 0.633 & 0.647 & 0.906 & 0.153 & 0.120 & 0.509 & 0.648 & \textbf{0.956} \\
    table & 0.233 & 0.287 & 0.772 & 0.080 & 0.095 & 0.577 & 0.307 & \textbf{0.880} & 0.442 & 0.462 & 0.886 & 0.195 & 0.221 & 0.743 & 0.494 & \textbf{0.963} \\
    chair & 0.382 & 0.433 & 0.802 & 0.050 & 0.088 & 0.447 & 0.481 & \textbf{0.875} & 0.617 & 0.615 & 0.913 & 0.134 & 0.345 & 0.665 & 0.693 & \textbf{0.964} \\
    sofa  & 0.499 & 0.535 & 0.786 & 0.058 & 0.129 & 0.577 & 0.638 & \textbf{0.895} & 0.725 & 0.708 & 0.895 & 0.153 & 0.208 & 0.734 & 0.834 & \textbf{0.972} \\
    \hline
  average & 0.407 & 0.446 & 0.769 & 0.062 & 0.091 & 0.468 & 0.425 & \textbf{0.872} & 0.618 & 0.626 & 0.885 & 0.158 & 0.209 & 0.647 & 0.649 & \textbf{0.959} \\
    \hline
    \end{tabular}%
    
    \medskip
    
    \begin{tabular}{c|cccccccc|ccccccc}
    \hline
    \multirow{2}[4]{*}{} & \multicolumn{8}{c|}{Chamfer Distance ($\times 100$) $\downarrow$}                 & \multicolumn{7}{c}{Normal Consistency $\uparrow$} \\ \hline
        \multicolumn{1}{c|}{category} & \tiny{PSR}   & \tiny{MC}    & \tiny{BPA}   & \tiny{ATLAS} & \tiny{DMC} & \tiny{DSDF}  & \tiny{DGP}   & \tiny{OURS}  & \tiny{PSR}   & \tiny{MC}    & \tiny{BPA}   & \tiny{ATLAS} & \tiny{DMC} & \tiny{DSDF}    & \tiny{OURS} \\ \hline
    display & 0.273 & 0.269 & 0.093 & 1.094 & 0.662 & 0.317 & 0.293 & \textbf{0.069} & 0.889 & 0.842 & 0.952 & 0.828 & 0.882 & 0.932 & \textbf{0.974} \\
    lamp  & 0.227 & 0.244 & 0.060 & 1.988 & 3.377 & 0.955 & 0.167 & \textbf{0.053} & 0.876 & 0.872 & 0.951 & 0.593 & 0.725 & 0.864 & \textbf{0.963} \\
    airplane & 0.217 & 0.171 & 0.059 & 1.011 & 2.205 & 1.043 & 0.200 & \textbf{0.049} & 0.848 & 0.835 & 0.926 & 0.737 & 0.716 & 0.872 & \textbf{0.955} \\
    cainet & 0.363 & 0.373 & 0.292 & 1.661 & 0.766 & 0.921 & 0.237 & \textbf{0.112} & 0.880 & 0.827 & 0.836 & 0.682 & 0.845 & 0.872 & \textbf{0.957} \\
    vessel & 0.254 & 0.228 & 0.078 & 0.997 & 2.487 & 1.254 & 0.199 & \textbf{0.061} & 0.861 & 0.831 & 0.917 & 0.671 & 0.706 & 0.841 & \textbf{0.953} \\
    table & 0.383 & 0.375 & 0.120 & 1.311 & 1.128 & 0.660 & 0.333 & \textbf{0.076} & 0.833 & 0.809 & 0.919 & 0.783 & 0.831 & 0.901 & \textbf{0.962} \\
    chair & 0.293 & 0.283 & 0.099 & 1.575 & 1.047 & 0.483 & 0.219 & \textbf{0.071} & 0.850 & 0.818 & 0.938 & 0.638 & 0.794 & 0.886 & \textbf{0.962} \\
    sofa  & 0.276 & 0.266 & 0.124 & 1.307 & 0.763 & 0.496 & 0.174 & \textbf{0.080} & 0.892 & 0.851 & 0.940 & 0.633 & 0.850 & 0.906 & \textbf{0.971} \\
    \hline
   average & 0.286 & 0.276 & 0.116 & 1.368 & 1.554 & 0.766 & 0.228 & \textbf{0.071} & 0.866 & 0.836 & 0.923 & 0.695 & 0.794 & 0.884 & \textbf{0.962} \\
    \hline
    \end{tabular}%
  \label{tab:quantitative}%
\end{table}%

We compare our method to both traditional surface reconstruction methods and learning-based mesh generative methods. Traditional methods include screened Poisson surface reconstruction (PSR)~\cite{kazhdan2006poisson,kazhdan2013screened}, marching cube (APSS variant)~\cite{lorensen1987marching,guennebaud2008dynamic}, and ball-pivoting algorithm (BPA)~\cite{bernardini1999ball}. Learning-based methods include AtlasNet~\cite{groueix2018papier}, Deep Geometric Prior (DGP)~\cite{williams2019deep}, Deep Marching Cubes (DMC)~\cite{liao2018deep}, and DeepSDF~\cite{park2019deepsdf}, which are representatives of different paradigms. Since meshes in ShapeNet are not manifolds, it's not trivial to calculate point normals with consistent directions, but many algorithms rely on correct normals. We thus employ PCPNet~\cite{guerrero2018pcpnet} to predict normals for the input point clouds. We then utilize MeshLab~\cite{cignoni2008meshlab} to reconstruct the meshes for the three traditional methods. Specifically, for Poisson surface reconstruction, an outlier removal as post-processing is applied. Since the ball-pivoting algorithm is sensitive to the radius, we tried the auto-guess mode of the MeshLab and also selected 3 radii manually to choose the best radius. Please refer to the supplementary materials for more details about the baseline algorithms.

We use  F-score~\cite{wang2018pixel2mesh},
Chamfer distance~\cite{fan2017point}, and normal consistency score~\cite{mescheder2019occupancy} to evaluate the methods. Specifically, we uniformly sample $10^6$ points on the reconstructed mesh and the ground truth mesh respectively, and calculate a normal for each point. For F-score, the precision and recall are calculated by checking the percentage of points in one point set that can find a neighbor from the other point set within a threshold $\mu$. The F-score is then calculated as the harmonic mean of precision and recall. To align with different sampling densities, $\mu$ is set to $\sqrt{S/10^6}$ for each shape and $S$ is the surface area of the ground truth mesh. For Chamfer Distance (CD), it measures the mean distance between each point in one point set to its nearest neighbor in the other point set. The normal consistency score is defined as the average of the absolute dot product between the normals in one mesh and the normals of the corresponding nearest points in the other mesh.

\begin{figure}[H]
\centering
\includegraphics[width=0.99\linewidth]{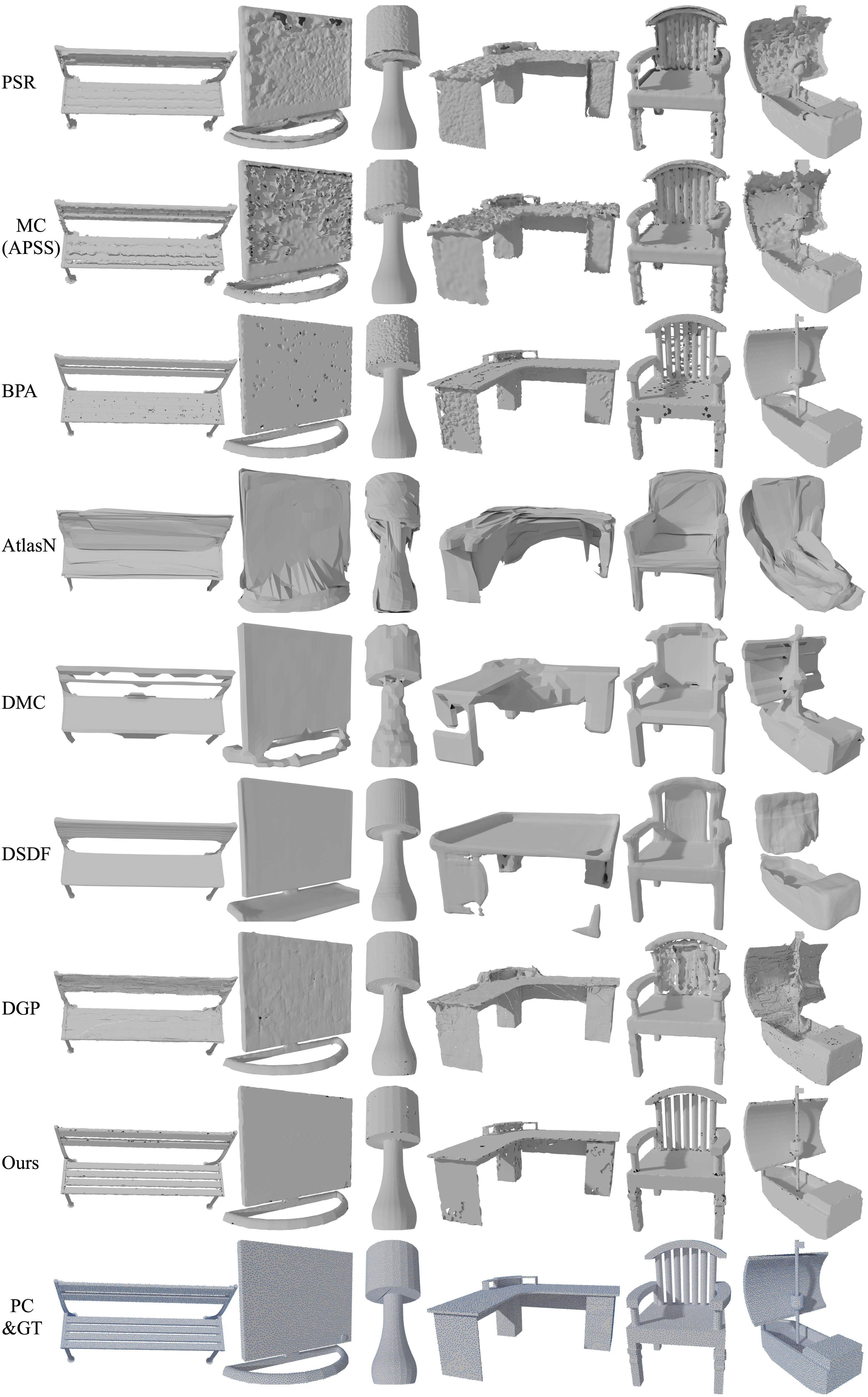}
\caption{Poisson surface reconstruction, marching cube (APSS), ball-pivoting algorithm, AtlasNet, Deep Marching Cubes, DeepSDF, Deep Geometric Prior, our method, and ground-truth meshes with input point clouds are shown from top to bottom.}
\label{fig:comparision}
\end{figure}

The quantitative results are shown in Table~\ref{tab:quantitative}. Our method outperforms all the baseline algorithms with regard to the three metrics across all categories with a large margin. Fig.~\ref{fig:comparision} demonstrates some of the representative cases. As shown in the figure, existing learning-based mesh generative methods (e.g., AtlasNet, DMC, and DeepSDF) are difficult to generate fine-grained details and generalize to unseen shapes (see the vessel of AtlasNet and the desk of DeepSDF). They even fail to preserve all the structures which are revealed in the input (see the bench and the desk of DMC for missing parts). This is due to the fact that they generate meshes based only on an object-level shape embedding. Also, since DMC utilizes 3D convolutional networks to predict the surface, the generated meshes are limited to a low resolution (i.e., $32\times32\times32$). As for the three traditional methods, they generally preserve the overall structures from the input point clouds. However, it may be difficult for them to handle ambiguous structures when the resolution of the input point cloud is limited. For example,  
\noindent they failed to distinguish the thin structures consisting of spatially close surfaces, such as the display and the desk, and produced much distortion. Without priors and reasoning about the shape, nor can they distinguish those independent but spatially adjacent parts, such as the long strips of the bench and the armchair. In contrast, our method fully utilizes the input point cloud, which enables the generation of fine-grained structures. The learned local priors also help us to better estimate the local connectivity and generalize to unseen categories. 

\subsection{Ablation Studies}
We would like to evaluate the transferability of our method, the importance of the candidate filtering, and the robustness with regard to various situations. 

\subsubsection{Category Transferability}
\begin{figure}[t]
    \centering
    \includegraphics[width=0.84\linewidth]{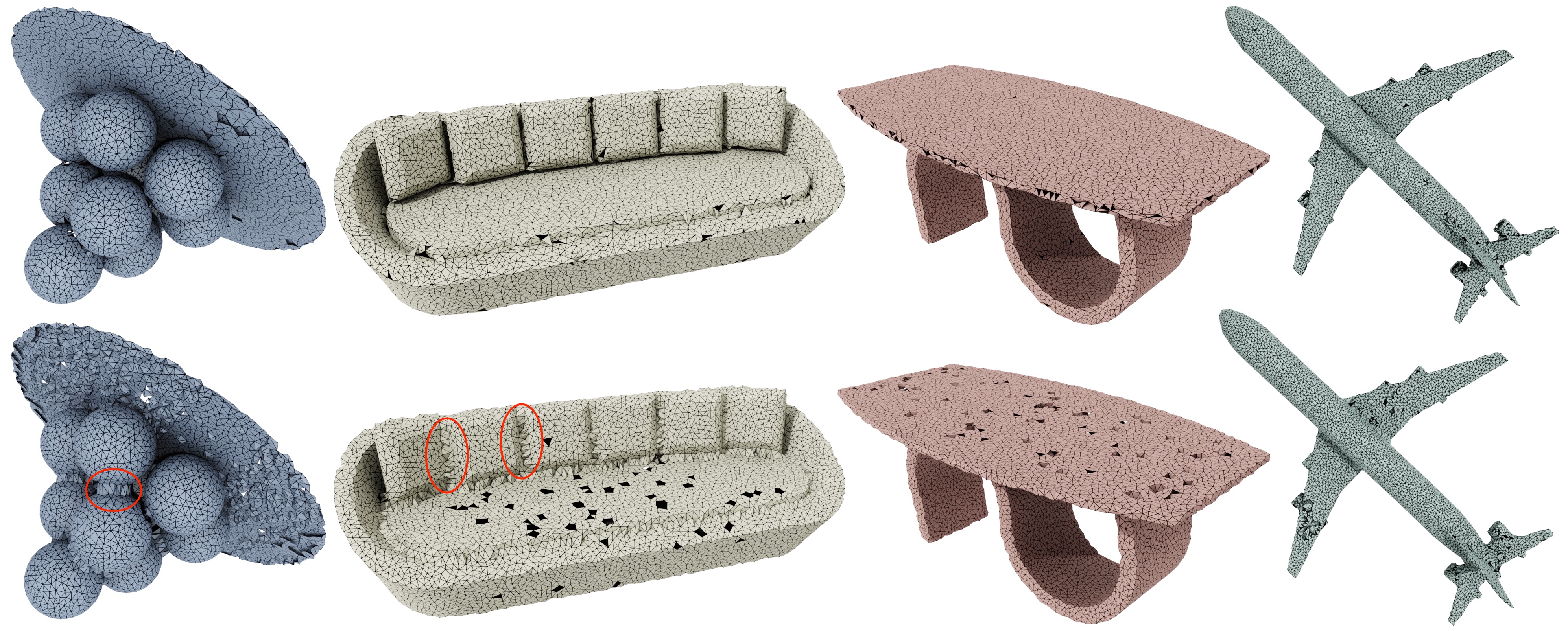}
    \caption{Our method can transfer to unseen categories. The first row shows the results of our method where the shape categories are unseen during training. The second row shows the results of ball-pivoting algorithm for comparison.}
    \label{fig:transfer}
\end{figure}

\begin{table}[t]
\centering
\captionof{table}{First row: the results of our leave-one-out cross-validation. Second row: the results of feeding all candidates directly to the post-processing algorithm without filtering. Last row: our original results for comparison.}
\scriptsize
\begin{tabular}{c|cccc}
\hline
& F-score ($\mu$) $\uparrow$ & F-score (2$\mu$) $\uparrow$ &CD($\times 100$) $\downarrow$ & normal similarity $\uparrow$ \\
\hline
leave-one-out & 0.870 & 0.959 & 0.072 & 0.961 \\
w/o filtering & 0.728 & 0.882 & 0.110 & 0.862  \\ \hline
ours & 0.872 & 0.959 & 0.071 & 0.962 \\
    \hline
\end{tabular}%
  \label{tab:ablation}%
\end{table}

We utilize leave-one-out cross-validation to evaluate the category transferability of our method. Specifically, we trained a separate model for each of the eight categories with the training data of all the rest seven categories, and a test is made for that category. Table~\ref{tab:ablation} reports the quantitative results where the numbers are averaged across all the eight categories. Compared to the model that trained on all categories and tests on all categories (the third row), the performances are quite similar, from which we can infer that our method does not heavily rely on category-specific priors and thus enable strong generalizability. Examples in Fig.~\ref{fig:transfer} have further confirmed our belief that local priors can be transferred across different categories.

\subsubsection{Effect of Filtering} 
\begin{figure}[t]
    \centering
    \includegraphics[width=0.8\linewidth]{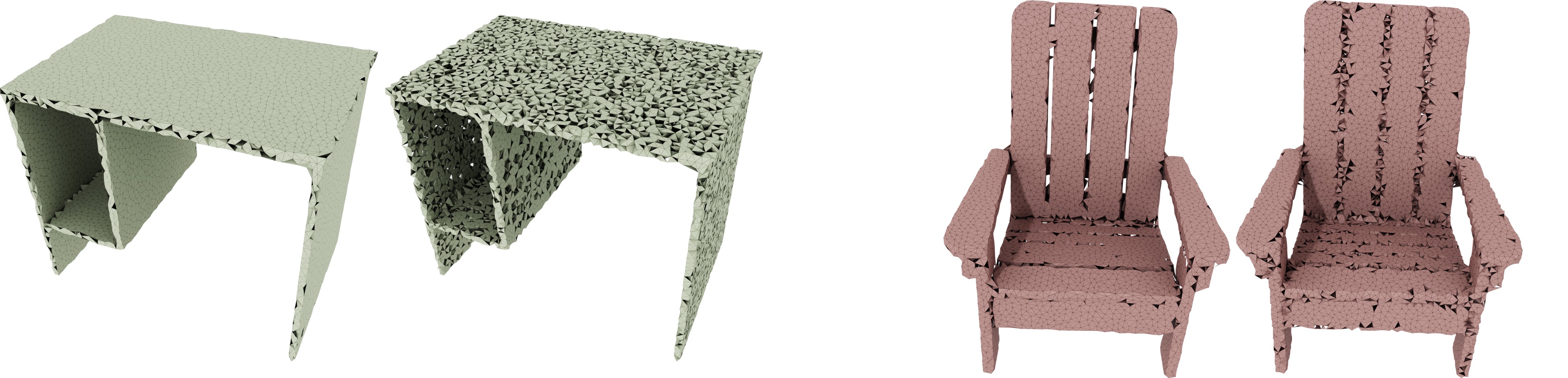}
    \caption{In each pair, the left one is the result of method, and the right one is the result without candidate filtering.}
    \label{fig:filtering}
\end{figure}

To verify the importance of the candidate filtering, we also test a variant where all the proposed candidates are directly fed into the post-processing algorithm without filtering. The quantitative results are shown in Table~\ref{tab:ablation}, from which we find that the performance drops dramatically. Fig.~\ref{fig:filtering} also shows some of the qualitative comparisons. Without passing the candidates through the network and filtering out all the incorrect candidates, the method cannot handle ambiguous structures anymore.

\subsubsection{Distribution of Point Cloud}
\begin{figure}[t]
    \centering
    \includegraphics[width=0.9\linewidth]{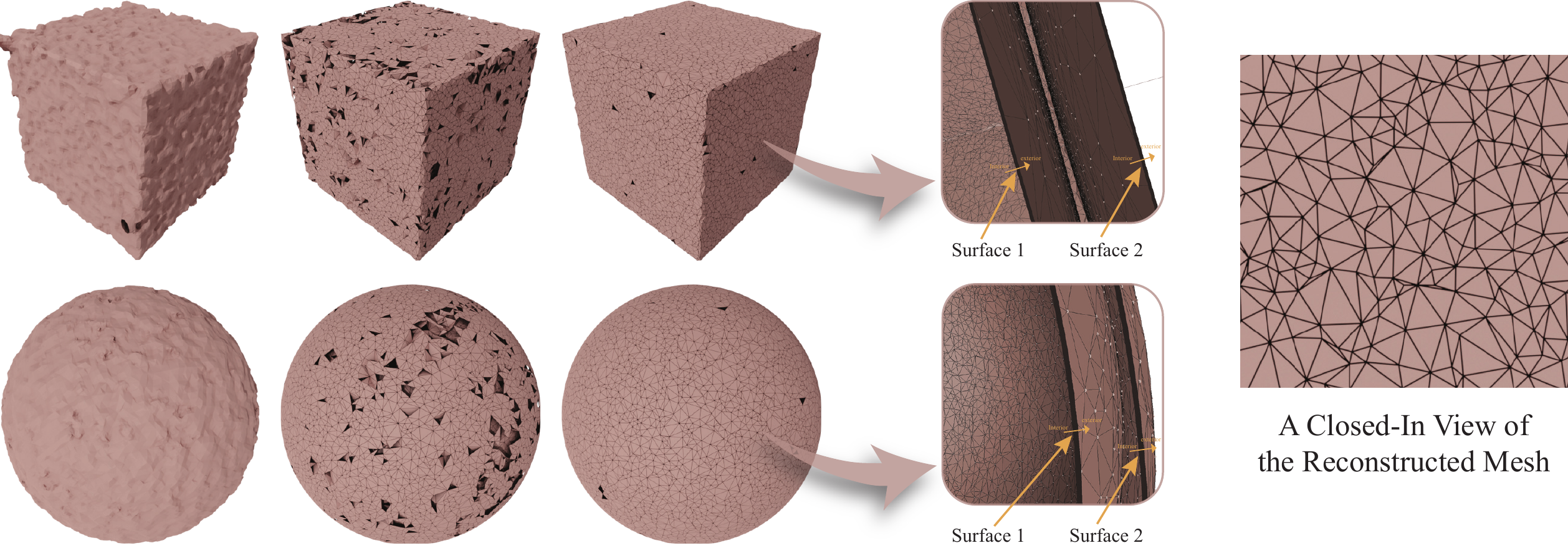}
    \caption{Qualitative results on uniformly randomly sampled point clouds. The ground truth meshes are concentric dual spheres and concentric dual cubes. From left to right: results from Poisson surface reconstruction (PSR)~\cite{kazhdan2006poisson,kazhdan2013screened}, results from the ball-pivoting algorithm~\cite{bernardini1999ball}, and results from our method. The second to the last column is a zoomed-in sliced view showing the interior of our results. The last column is a zoomed-in view of the reconstructed surface.}
    \label{fig:juniform}
\end{figure} 
In general, our method favors evenly distributed point clouds, and applying a Poisson-disk sampling as pre-processing could improve the performance. To examine the robustness to other point cloud distributions, we test our method on uniformly randomly sampled point clouds, virtual scanned point clouds, as well as Poisson-disk sampling with different density. Due to the space limits, we only include the results of uniformly randomly sampled point clouds in the text. Please refer to the supplementary materials for the results of other experiments. Note that though misleading, the uniformly randomly sampled points are typically not evenly distributed over the surface, as shown in the last column of Fig.~\ref{fig:juniform}. Two simple shapes are tested, namely the concentric dual spheres and concentric dual cubes. It can be seen from Fig.~\ref{fig:juniform} that although we only trained on evenly distributed Poisson-disk sampled point clouds (on ShapeNet), our method can process the uniformly randomly sampled point clouds effectively, and outperforms both PSR and BPA by a large margin. This further proves the strong generalizability of our method.

\subsubsection{Noisy Data and Real Scans} 

\begin{figure}[t]
\centering
\includegraphics[width=0.8\linewidth]{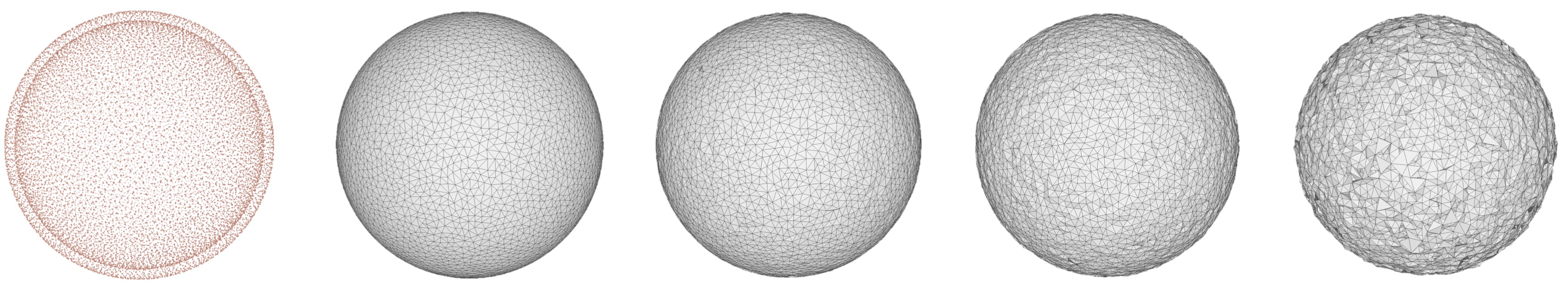}
\includegraphics[width=0.8\linewidth]{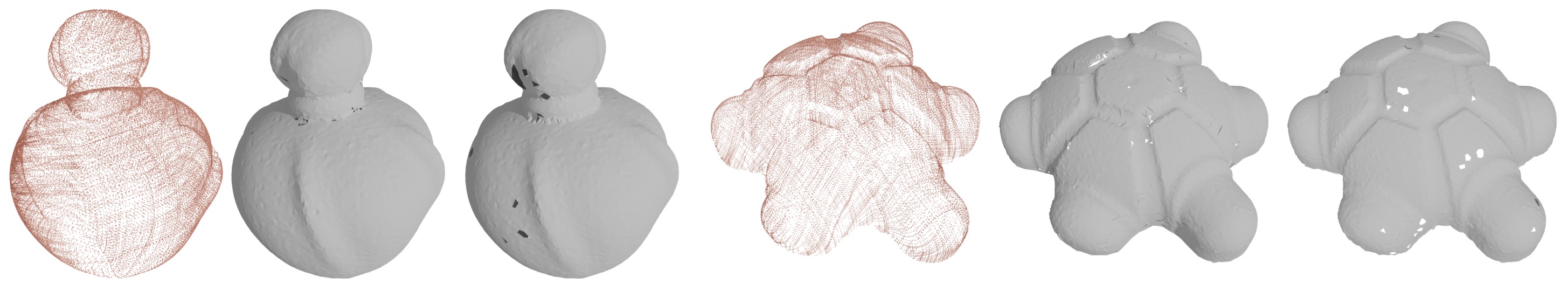}
\caption{First row: a point cloud of a concentric dual sphere and reconstructed meshes with different levels of input noise. Second row: two real-world LiDAR scans from Aim@Shape and the reconstructed meshes by our method.} 
\label{fig:noisy}
\end{figure}

Since our method directly interpolate triangles upon input point clouds, the algorithm may be sensitive to the noise. However, as shown in Fig.~\ref{fig:noisy}, without explicit denoising and data augmentation mechanisms, our method is still resistant to the noise to a certain extent. As for real scans, there may be more issues, such as part missing and uneven distribution of the points. With the point set consolidation network~\cite{yu2018ec} as pre-processing, our method can generate satisfying meshes from real-world LiDAR scans (see Fig.~\ref{fig:noisy}). In the future, we would like to explore explicit ways to propose the position of the vertices and compensate for the structural loss of input~\cite{liu2019morphing}.

\section{Conclusions}

In this paper, we proposed a novel learning-based framework for mesh reconstruction that is based on the grounded point clouds and explicitly estimates the local connectivity of the points. By leveraging the intrinsic-extrinsic ratio as training guidance, the method is able to effectively distinguish the surface triangles and non-surface triangles. Extensive experiments have shown our superior performance, especially for preserving the details, handling ambiguous structures, and strong generalizability.

\subsubsection{Acknowledgements} This work was funded in part by Kuaishou Technology, NSF grant IIS-1764078, NSF grant 1703957, the Ronald L. Graham chair and the UC San Diego Center for Visual Computing.
\clearpage
%
%
\bibliographystyle{splncs04}
\bibliography{egbib.bib}

\newpage

\section*{Appendix A: More figures of the Reconstructed Meshes}
\begin{figure}[H]
\centering
\vskip -2em
\includegraphics[width=\linewidth]{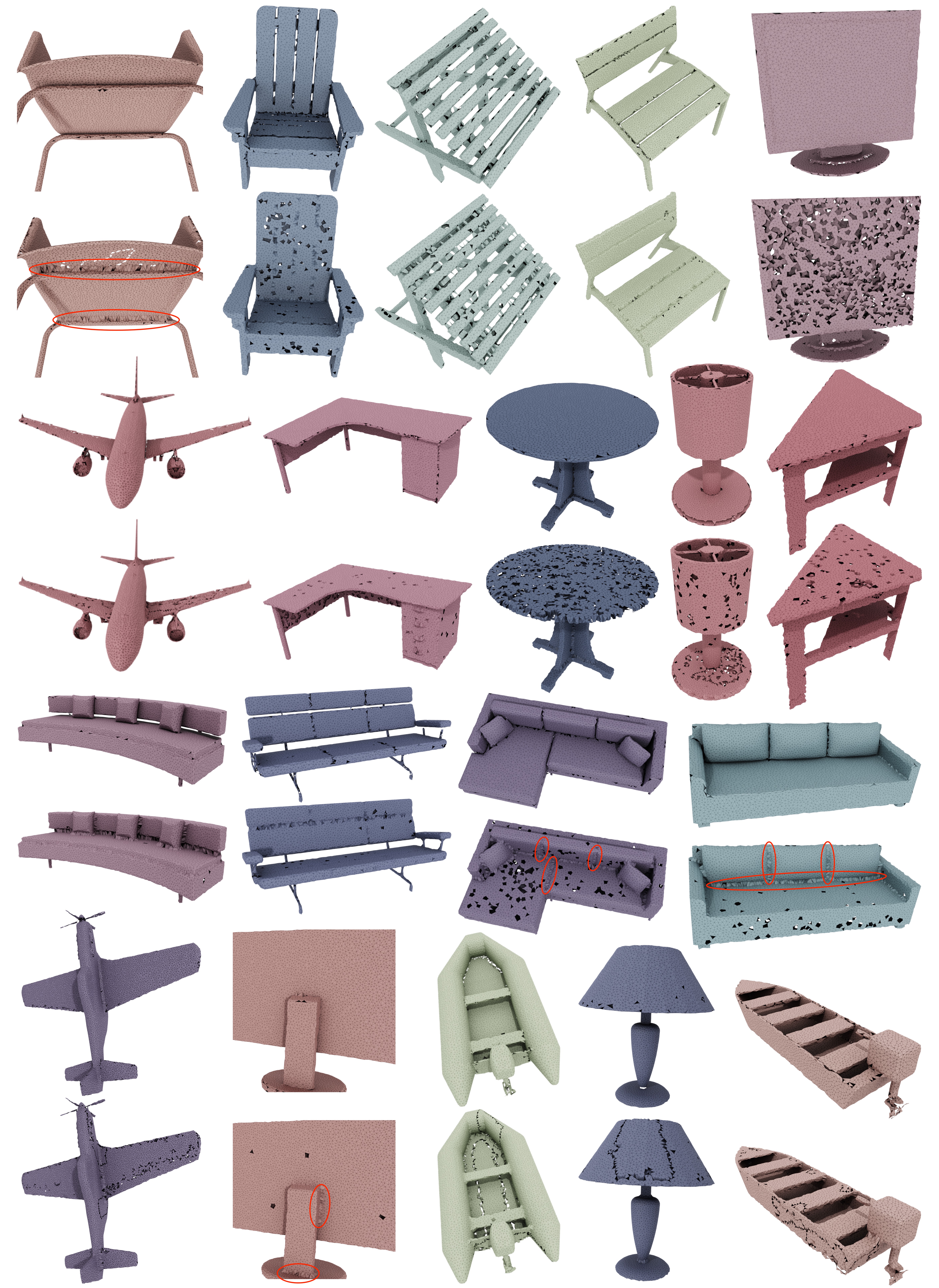}
\caption{Reconstructed meshes of the ShapeNet test set. In each pair, the above one is the result of our method, and the below one is the result of the traditional ball-pivoting algorithm.} 
\label{fig:noisy}
\end{figure}

\section*{Appendix B: Pseudo Code of the Remeshing Algorithm}

 \begin{algorithm}[H]
 \SetAlgoLined
 \SetKwInOut{Input}{input}
 \SetKwInOut{Output}{output}
 \SetKw{Continue}{continue}
 \Input{Reference mesh $M_R$, point cloud $P$ sampled on $M_R$, IER threshold $\tau$}
 $M_N.V = P$\;
 $M_N.F = \emptyset$\;
 Construct a $k$-NN graph on $P$ and propose candidate triangle faces\;
 Calculate the intrinsic-extrinsic ratio for each candidate\;
 Filter out the incorrect candidates with IER $\geq \tau$ \;
 Sort the remaining candidates with respect to their distance to $M_R$ and the length of their longest edges\;
 \For{each remaining candidate triangle $f_i$} 
 {
 \If{$\exists\ f_j \in M_N.F$ intersects with $f_i$ \textbf{or} $\exists\ edge \in M_N$  has more than two incident faces after adding $f_i$ }
 {\Continue;}
 \Else{$M_N.F = M_N.F \cup \{f_i\}$\;}
 }
\Return{$M_N$}\;
  \caption{Remeshing with Intrinsic-Extrinsic Ratio as Guidance} 
  \label{alg:remeshing}
 \end{algorithm}

\section*{Appendix C: Confusion Matrix of the Candidate Classification}

 Table~\ref{tab:confuse matrix} shows the confusion matrix of the candidate classification on the ShapeNet test set. Specifically, category 0 indicates the incorrect triangles filtered out by the IER. Both category 1 and 2 are the correct candidates. The candidates of category 1 are closer to the ground truth surface than candidates of category 2. We find that the overall performance of the candidate classification is satisfactory.
 \begin{table}[H]
 \centering
 \vskip -2em
 \caption{Confusion matrix of the candidate classification on the ShapeNet test set. }
 \begin{tabular}{c|ccc}
 \hline
 & category 0 & category 1 & category 2 \\ \hline
 category 0 & 89.8$\%$  & 3.9 $\%$ & 6.3$\%$ \\
 category 1 & 1.4$\%$  & 96.5$\%$  & 2.1$\%$ \\
 category 2 & 20.2$\%$  & 8.8$\%$  & 71.0$\%$ \\
     \hline
 \end{tabular}%
   \label{tab:confuse matrix}%
 \end{table} 
 
\section*{Appendix D: Results of Different Sampling Densities}
\begin{figure}[H]
 \centering
 \vskip -2em
 \includegraphics[width=\linewidth]{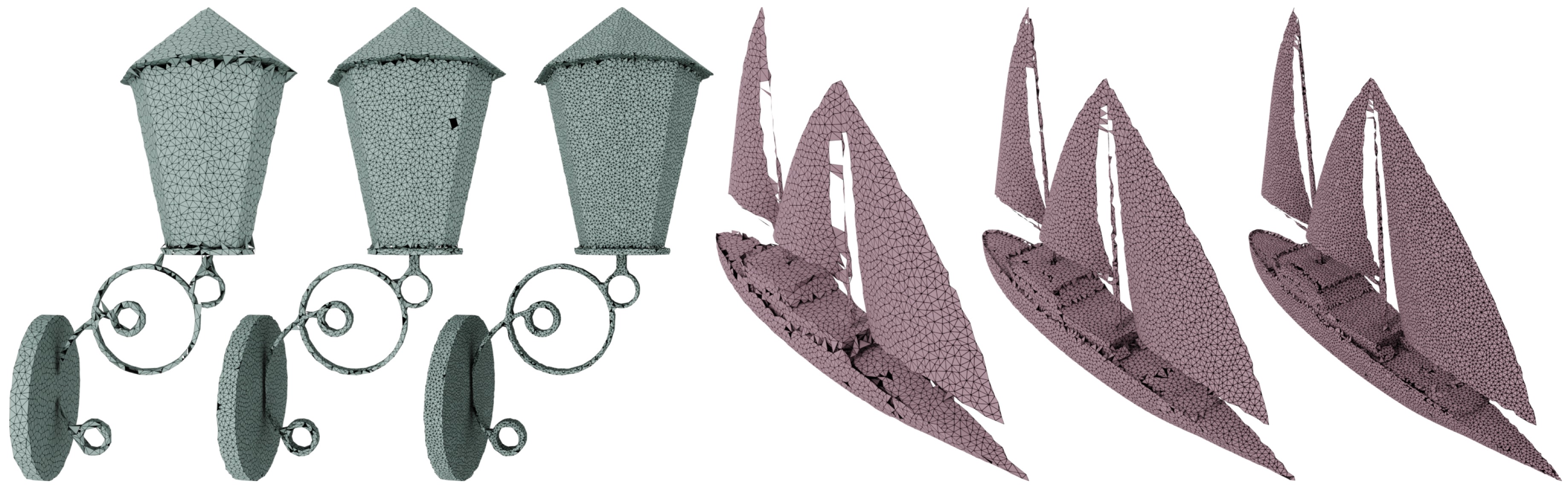}
 \caption{Qualitative results of different sampling densities. Training on point clouds with 12,800 points (middle), our method can transfer to point clouds with 6,400 (left) and 25,600 points (right).}
 \label{fig:density}
 \end{figure}
 
 \begin{table}[H]
 \vskip -4em
 \centering
 \caption{Quantitative results on point clouds with different densities (F-score with two thresholds, Chamfer distance, and normal consistency score). The results are averaged across the eight categories.}
 \begin{tabular}{c|cccc}
 \hline
\#points & F-score($\mu$) $\uparrow$ & F-score($2\mu$) $\uparrow$ & CD ($\times 100$)  $\downarrow$   & normal $\uparrow$ \\
 \hline
     6,400 & 0.814 & 0.916 & 0.091 & 0.949 \\
     12,800 & 0.872 & 0.959 & 0.071 & 0.962 \\
     25,600 & 0.907 & 0.983 & 0.062 & 0.969 \\
 \hline
 \end{tabular}%
   \label{tab:density}%
 \end{table}
 
 To evaluate the transferability of our method across different sampling densities. We trained our models on point clouds with $12,800$ points and test it on point clouds with $6,400$ and $25,600$ points respectively. Fig.~\ref{fig:density} shows the qualitative results, from which we find that our method can generalize to different density distributions, and as the resolution of the point clouds increases, the details become more accurate. The quantitative results shown in Table~\ref{tab:density} further confirm our arguments.
 
 \section*{Appendix E: Results on Virtual Scans}
 
 In order to examine the robustness of our method with regard to different distributions of the input point clouds, we test our method on virtual scans of the ShapeNet models. Specifically, we utilize the VCG Lib to mimic the Kinect sensors and randomly select 10 camera poses to scan the models. The scanned point clouds of different views are fused and downsampled to 12,800 points (Poisson-disk sampling) before feeding into our method. 
 
\begin{figure}[H]
 \centering
 \vskip -1em
 \includegraphics[width=\linewidth]{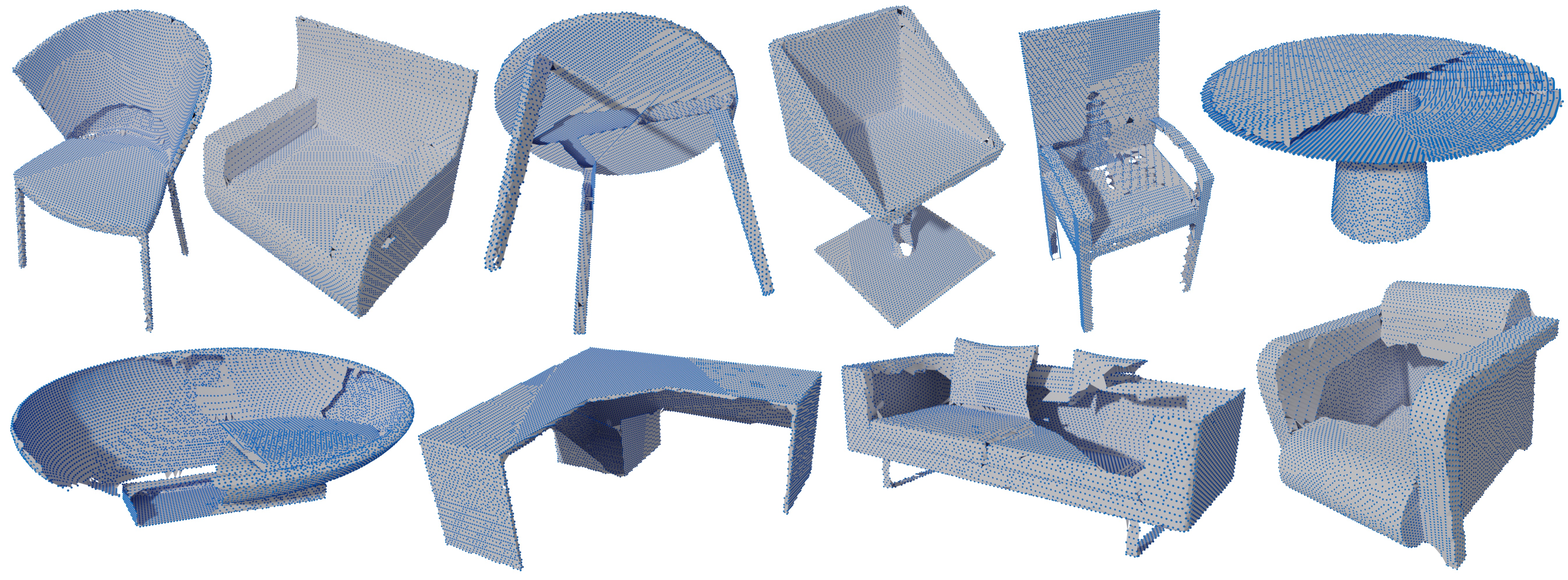}
 \caption{Input point clouds with 12,800 points and the reconstructed meshes by our method.}
 \label{fig:virtual_scan}
 \end{figure}

 Fig.~\ref{fig:virtual_scan} shows the point clouds and the reconstructed meshes. Although the scanned points clouds are unevenly distributed, our method still reconstructs high-quality meshes, which demonstrates the generalizability of our method. Please note that due to the limited number of views, the scanned point clouds may not cover the full area of the shape and the reconstructed meshes are thus also incomplete. 

\section*{Appendix F: Hyper-parameters of the Experiments}

Some baseline algorithms require the normals of the point clouds as input. However, meshes in ShapeNet are not perfect manifolds, there are lots of flipped faces and faces that are visible from multiple views. As a result, it's not trivial to determine the consistent directions of the normals (point inward or point outward). We thus employ PCPNet to predict normals for the input point clouds. We then utilize MeshLab to reconstruct the meshes for the three traditional methods. We basically follow the default hyper-parameters of MeshLab. Specifically, the reconstruction depth of PSR is set to be 8, and the grid resolution of Marching Cube (APSS) is set to be 200. Since ball-pivoting algorithms are sensitive to the radius, we tried the auto-guess mode of the MeshLab and also manually selected 3 radii $1\%, 2\%,$ and $3\%$ to choose the best radius. For PSR, we also provide outlier removal as post-processing, which filters out all the vertices that cannot find a point in the input point cloud within a radius of 0.02. 

For all the learning-based methods, we used our training set (point clouds with 12,800 points) and followed their released hyperparameters to retrain the model. For DeepSDF, we retrained the network category-by-category. Since both positive and negative signed distance samples are required by the DeepSDF, we assume the normal direction is known for each point in order to sample testing signed distance samples. For AtlasNet, we use the ``Autoencoder 25 Squares'' model. For Deep Geometric Prior (DGP), ``radius'' is set to be 0.05, ``local-epochs'' and ``global-epochs'' is set to be 125, and ``upsamples-per-patch'' is set to be 64. Since DGP outputs point clouds of millions of points and reconstructing meshes on such point clouds is time-consuming, we directly use the generated points to calculate the F-score and Chamfer distance, and do not report the normal consistency score.

For the noise experiments, we test on the point clouds of concentric dual spheres. The diameters of the two spheres are 0.933 and 1 respectively. We add a Gaussian noise of a standard deviation of $0.001\times t$ to point coordinates, where $t$ indicates the level of the noise. For the figure of the main paper, $t$ is set to be 0, 0.8, 1.6, and 3.2 respectively.

\section*{Appendix G: Geodesic Distance Regression}
In our method, we use the ratio of the geodesic distance and Euclidean distance as guidance to train the network to classify the candidate triangles. Another straightforward idea is that regressing the geodesic distances between pairs of vertices directly with methods such as GeoNet and then use the regressed geodesic distance to classify the candidates. We tried this ablated version to estimate its effectiveness. Specifically, since GeoNet didn't release their source code, we modify our classification network to regress the geodesic distance directly. As we only care about the geodesic distance within a small neighborhood, the training set only contains the pairs between each vertex and its $k$-nearest neighbors, and the distances are truncated with a threshold of 0.1. 

\begin{table}[H]
  \vskip -2em
  \centering
  \caption{Results of the geodesic distance regression. The first row shows the relative error of the predicted distance. The second row shows the accuracy of the candidate classification using the predicted geodesic distance.}
    \begin{tabular}{l|rrrrrrrr|r}
    \hline
    category      & \multicolumn{1}{l}{airplane} & \multicolumn{1}{l}{cabinet} & \multicolumn{1}{l}{chair} & \multicolumn{1}{l}{display} & \multicolumn{1}{l}{lamp} & \multicolumn{1}{l}{sofa} & \multicolumn{1}{l}{table} & \multicolumn{1}{l}{vessel} & \multicolumn{1}{|l}{average} \\
    \hline
    relative error & 137.7\% & 44.0\% & 59.8\% & 47.6\% & 132.0\% & 47.5\% & 49.1\% & 91.7\% & 70.7\% \\
    accuracy & 58.0\% & 83.2\% & 61.9\% & 70.1\% & 47.4\% & 71.1\% & 73.3\% & 51.6\% & 65.7\% \\
    \hline
    \end{tabular}%
  \label{tab:regress}%
\end{table}%

Table~\ref{tab:regress} shows the results of our regression version. We find that it's not easy for the network to regress the geodesic distances and the predicted distances are not accurate. Using the predicted geodesic distance to classify the candidate triangles (into 2 categories) also produces poor results. The accuracy of $65.7\%$ is much lower than the accuracy of our original version of $91.8\%$. In fact, in our original version, the network does not need to regress the geodesic distance. The labels inferred by the ratio of the geodesic distance and the Euclidean distance only serve as the supervision for the network to learn some local priors to recognize those incorrect triangles, which is a more reasonable and easy task. 
\end{document}